\pdfoutput=1

\documentclass[11pt]{article}

\usepackage[final]{acl}

\usepackage[T1]{fontenc}
\usepackage[utf8]{inputenc}

\usepackage{times}
\usepackage{latexsym}
\usepackage{amsmath}
\usepackage{amsfonts}
\usepackage{array}

\usepackage{subcaption}  
\usepackage{microtype}
\usepackage{inconsolata}
\usepackage{graphicx}
\usepackage{color}
\usepackage{float}

\title{
When Annotators Disagree, Topology Explains: Mapper, a Topological Tool for Exploring Text Embedding Geometry and Ambiguity}

\author{
 \textbf{Nisrine Rair\textsuperscript{1,2}},
 \textbf{Alban Goupil\textsuperscript{1}},
 \textbf{Valeriu Vrabie\textsuperscript{1}},
 \textbf{Emmanuel Chochoy\textsuperscript{2}}
\\
\\
 \textsuperscript{1}CReSTIC, Université de Reims Champagne-Ardenne, Reims, France,\\
 \textsuperscript{2}Chochoy Conseil, Reims, France
\\
\href{mailto:nisrine.rair@univ-reims.fr}{\texttt{nisrine.rair@univ-reims.fr}}
\\
}

\begin{document}
\maketitle

\begin{abstract} 
    Language models are often evaluated with scalar metrics like accuracy, but such measures fail to capture how models internally represent ambiguity, especially when human annotators disagree. We propose a topological perspective to analyze how fine-tuned models encode ambiguity and more generally instances.
    Applied to RoBERTa-Large on the MD-Offense dataset, Mapper, a tool from topological data analysis, reveals that fine-tuning restructures embedding space into modular, non-convex regions aligned with model predictions, even for highly ambiguous cases. Over $98\%$ of connected components exhibit $\geq 90\%$ prediction purity, yet alignment with ground-truth labels drops in ambiguous data, surfacing a hidden tension between structural confidence and label uncertainty.
    Unlike traditional tools such as PCA or UMAP, Mapper captures this geometry directly uncovering decision regions, boundary collapses, and overconfident clusters. Our findings position Mapper as a powerful diagnostic tool for understanding how models resolve ambiguity. Beyond visualization, it also enables topological metrics that may inform proactive modeling strategies in subjective NLP tasks. For reproducibility, all code and experiment configurations are released\footnote{\url{https://github.com/NisrineRair/tda-nlp-representations}}.
\end{abstract}

\section{Introduction}

Ambiguity is a persistent challenge in Natural Language Processing (NLP) \citep{pavlick_inherent_2019} when instances often permit multiple plausible meanings or evoke differing judgements across annotators \citep{wan_everyones_2023}. This is the case in offensive language detection or natural language inference, where annotators may disagree \citep{leonardelli_agreeing_2021, nie_what_2020}. While prior work has focused on modeling the ambiguity using soft labels or confidence estimates \citep{swayamdipta_dataset_2020, fornaciari_beyond_2021}, less is known about how ambiguous examples are internally structured in representation space.

\begin{figure}
    \centering
    \begin{subfigure}{0.45\columnwidth}
        \includegraphics[width=\textwidth]{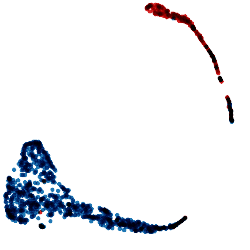}
        \caption{UMAP projection} 
        \label{fig:GeometryVsTopologyA}
    \end{subfigure} 
    \hfill
    \begin{subfigure}{0.45\columnwidth}
        \includegraphics[width=\textwidth]{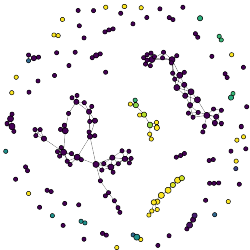}
        \caption{Mapper graph}
        \label{fig:GeometryVsTopologyB}
    \end{subfigure}
    \caption{Illustration of geometric (UMAP) and topological (Mapper) views of model representations.  Figure(\subref{fig:GeometryVsTopologyA}) UMAP shows apparent separation between \textcolor[RGB]{220,20,60}{offensive}  and \textcolor[RGB]{65,105,225}{non-offensive} instances. (\subref{fig:GeometryVsTopologyB}) Mapper reveals a more fragmented and connected structure, with \textcolor[RGB]{255,215,0}{offensive} and \textcolor[RGB]{72,61,139}{non-offensive} regions interwoven, highlighting spatial organization that UMAP flattens.}
    \label{fig:GeometryVsTopology}
\end{figure}

Fine-tuning of language models is known to rearrange data within the embedding spaces, increasing cohesion and alignment with task-relevant categories \citep{zhou_directprobe_2021, rajaee_how_2021}. These changes are often analyzed using geometric tools such as PCA and UMAP, and quantified using  metrics like cosine similarity, silhouette score, and anisotropy \citep{ethayarajh_how_2019, coenen_visualizing_2019, cai_isotropy_2021}. Yet such tools can oversimplify high-dimensional structure. PCA's global projections compress variance into a limited number of axes \citep{wold_principal_1987}, while UMAP enforces local Euclidean neighborhoods \citep{mcinnes_umap_2020}, often at the expense of capturing nonlinear or hierarchical relationships. In contrast, linguistic embedding space has been shown to exhibit natural hyperbolic geometry \cite{nickel_poincare_2017}, a structure that topological methods can preserve without relying on Euclidean assumptions. This is particularly important in high-dimensional settings, as dimensionality increases, points tend to become nearly equidistant, blurring the notion of proximity and undermining the interpretability of distance-based measures \citep{donoho_high-dimensional_2000, goos_surprising_2001}. As a result, familiar geometric intuitions about neighborhoods and separation begin to break down, making it easy to misread the structure of the embedding space.
 
This limitation is illustrated in Figure~\ref{fig:GeometryVsTopologyA} where UMAP applied to the final-layer \texttt{[CLS]} embeddings of a RoBERTa-Large model fine-tuned on binary offensive language detection, presents the offensive class as a single, compact cluster. On the contrary, topological tools construct graphs based on point connectivity rather than distances, revealing a more fragmented and nuanced organization as in Figure~\ref{fig:GeometryVsTopologyB}, where offensive instances are spread across multiple distinct yet connected regions, suggesting that geometric projections may flatten meaningful topological structure. This example highlights a key limitation of classical tools: visual simplicity can come at the cost of structural fidelity. This limitation reflects the embedding space's non-Euclidean, hyperbolic structure.

In this work, we use Mapper, a topological tool, to analyze how fine-tuned language models embed ambiguous instances and how this structure manifests in the geometry of embedding space. Focusing on offensive language detection, we explore how fine-tuning reshapes topological organization, why models exhibit overconfidence in ambiguous inputs, and how these behaviors relate to human disagreement.
Our results show that fine-tuning imposes topological regularity: even instances with annotation disagreements are mapped into smooth, consistent regions. Rather than ignoring uncertainty, language models appear to spatially reorganize it, resolving ambiguity into reliable structures.

\section{Related Work}

\paragraph{Ambiguity, Disagreement, and Uncertainty in NLP.} Ambiguity in NLP has been studied from multiple angles. Early work often treated disagreement as label noise to be filtered~\citep{zhang_understanding_2017}, whereas more recent studies argue that it reflects genuine linguistic uncertainty and advocate for distributional supervision to capture annotator variability~\citep{pavlick_inherent_2019, mostafazadeh_davani_dealing_2022}. Building on this perspective, several studies investigated model behavior on ambiguous inputs. \citep{swayamdipta_dataset_2020} show that ambiguous examples exhibit unstable training dynamics, while \citet{liu_were_2023} argue that models tend to be overconfident on such inputs, failing to reflect underlying uncertainty in their predictions.
Other research has challenged the use of majority vote in subjective tasks. \citet{leonardelli_semeval-2023_2023} argue that majority aggregation suppresses meaningful disagreement signals, motivating soft-label and annotator aware modeling. These approaches assess ambiguity at the output level, leaving open the question of how it is structurally encoded within the model.

\paragraph{Fine-Tuning and Embedding Space.} 

Fine-tuning is known to reshape representation in the embedding space. Studies using distance-based metrics show that fine-tuned models for classification tasks increase intra-class cohesion and expand inter-class separation \citep{zhou_directprobe_2021}. Spectral analyses reveal that variance becomes concentrated in relevant directions, reducing isotropy of embedded instances \citep{cai_isotropy_2021, ethayarajh_how_2019}. Probing classifiers further confirm that linguistic features become more linearly separable after fine-tuning \citep{hewitt_structural_2019}.
However, distance-based metrics oversimplify or distort high dimensional structure, potentially missing connectivity patterns and latent organization.

\paragraph{Topological Methods in NLP.}

Topological Data Analysis (TDA) provides powerful tools for studying high-dimensional data beyond geometry, focusing on how points connect rather than how far apart they are. Two key tools are \textbf{(1) Mapper} \citep{singh2007topological}, which constructs graph networks by projecting data through a lens into lower dimensions, and \textbf{(2) Persistent homology}  \citep{carlsson_topology_2009}, which quantifies topological features like clusters or holes that persist across scales. While both methods have demonstrated significant success in different applications, for example in identifying novel cancer subtypes from gene expression data \citep{lum_extracting_2013}, their application to NLP remains limited, with few applications that can be categorized into three main directions:
\textbf{(1) Lexical-level analysis}, \citet{rathore_topobert_2023} employed Mapper to track topological transformations in word embedding spaces during fine-tuning, revealing how semantic relationships reorganize at different training stages.
\textbf{(2) Summarizing documents}, \citep{guan_topological_2016} demonstrated that homological simplification can remove noise while preserving structural patterns and effectively extract keyphrases by preserving the topological structure of document semantic graphs using TF-IDF weighted representations. 
\textbf{(3) Model-internal analysis}, \citet{fitz_hidden_2024} applied persistent homology to compare the topological complexity of LSTM and Transformer activation patterns, finding recurrent architectures exhibit greater representational redundancy. 
Complementary, \citet{gardinazzi_persistent_2024} utilized zigzag persistence to track the evolution of topological features across transformer layers, enabling principled model pruning based on structural redundancy analysis. While these prior works demonstrate the feasibility of topological tools in NLP, they focus largely on pretrained models, lexical structure, or model internals.

Prior work offers valuable insights into how models handle ambiguous data, how fine-tuning reshapes representation spaces, and how topological methods extract structures in high dimensions. However, these strands remain disconnected. Ambiguity is typically evaluated at the output level, fine-tuning effects are described through geometric metrics, and TDA is rarely applied beyond lexical or pretrained settings. Our work bridges these gaps by using Mapper TDA-algorithm to explain how fine-tuned models internally structure ambiguous instances revealing predictive regularities that persist even for annotator disagreement.

\section{The Mapper Algorithm}

\begin{figure*}
    \centering
    \includegraphics[width=\textwidth]{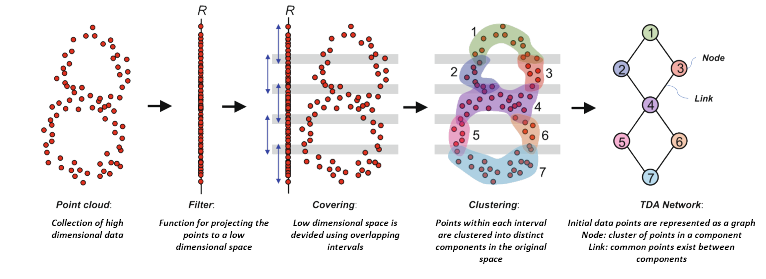} 
    \caption{Illustration of the Mapper algorithm. A high-dimensional data is first projected using a lens function (filter) to a low-dimensional space. The projection is covered by overlapping hypercubes (intervals here), and each preimage is clustered in the original space. Nodes in the resulting graph represent clusters, and edges indicate shared points due to overlap. Adapted  from Figure 1 of \citet{liao_tmap_2019}, available under CC BY 4.0.}
    \label{fig:mapper_pipeline}
\end{figure*}

This section provides a brief overview of the Mapper algorithm, the foundation of our topological analysis. Mapper~\citep{singh2007topological} constructs a graph-based summary of high-dimensional data by combining filtering, covering, and clustering. This process reveals global patterns while preserving local structure, offering a topological perspective on how data are organized. The algorithm steps, illustrated in Figure~\ref{fig:mapper_pipeline}, are described below.
\\
    {\bf Filtering.} Because high-dimensional data can be viewed from many angles, a \emph{lens function} allows it to be projected to highlight a specific aspect of interest.
    Formally, a lens is a function \( f : \mathbb R^d \to \mathbb R^k \), mapping each point in the original \( d \)-dimensional space to a \( k \)-dimensional space, where \( k \ll d \). While any value of \( k \) can be used, implementations typically use $k\le3$ to balance computational efficiency with interpretability. The lens function defines a coordinate system by which the data can be more easily partitioned in the next step. This flexibility allows Mapper to be tailored to different analysis goals. For example, projecting onto one or several principal components emphasizes directions of high variance, while ignoring details attached on the other directions. The lens can be any user-defined function, including geometric descriptors such as eccentricity or centrality, density-based scores, or task-specific measures, depending on what aspect of the data one aims to reveal.
    \\
     {\bf Covering.} The codomain, i.e. the image \( \mathbb{R}^k \) of the lens \( f \), is then covered by overlapping hypercubes. 
     Mapper divides each dimension into \( r \) equal-length intervals, forming a grid of hypercubes that partition the image of $f$. This grid defines a covering \( \mathcal{U} = \{U_i\} \), where each \( U_i \) is a hypercube of side length \( 1/r \). For 1D lens, the covering consists of intervals, for \( k = 2 \) it consists of rectangles, and in general, for arbitrary \( k \), the covering is a collection of overlapping hypercubes in \( \mathbb{R}^k \). The parameter \( r \) controls the granularity of this partition: higher values reveal local structure, while lower values favor generalization. To encourage continuity between hypercubes, Mapper introduces a fractional overlap \( \epsilon \in [0, 1) \) between adjacent bins.
     This expands each hypercube slightly by an amount \( \epsilon \cdot \frac{1}{r} \) in each dimension, so that neighboring hypercubes intersect. A point \( f(x) \) may therefore fall into multiple overlapping bins, and the corresponding data point \( x \) will be included in multiple preimage sets \( f^{-1}(U_i) \).  This overlap is essential for connectivity, since it facilitates the linking of topologically related clusters, preserving continuity and coherence of the underlying data structure in the final graph. 
    \\
    {\bf Clustering.} For each hypercube \( U_i \in \mathcal{U} \), Mapper identifies the corresponding subset \( f^{-1}(U_i) \cap X \) in the original high-dimensional space \( \mathbb{R}^d \)  and applies a clustering algorithm. Each cluster corresponds to a node in the resulting Mapper graph. This local clustering avoids reliance on global geometry and is well-suited to nonlinear data. The clustering algorithm can be chosen based on the task. Common choices include DBSCAN, HDBSCAN, or other density or distance-based methods. If a data point appears in multiple overlapping bins and is assigned to several clusters, the corresponding nodes in the Mapper graph are connected by an edge. These edges reflect shared membership across clusters and allow Mapper to summarize the data as a graph that captures both local structure and global connectivity. 
    \\
     {\bf Graph construction.} The final output of Mapper is a graph where each node corresponds to a cluster and edges connect nodes that share one or more data points. 
     A graph may contain multiple connected components, reflecting distinct regions of the dataset that remain disconnected under the lens and clustering choices. Analyzing the size, homogeneity, or connectivity of these components can reveal meaningful structures in the data.\\
Mapper's output depends on three key design choices: the lens function, which determines the aspect of the data being emphasized, the cover resolution and overlap, which control how finely the space is segmented and how regions connect, and the clustering algorithm, which determines how data points are grouped into nodes. Each of these elements shapes the granularity, connectivity, and interpretability of the resulting graph. Choosing values that balance structural detail with readability is essential. For a broader comparison of these elements and their impact, see \citet{madukpe_comprehensive_2025}.

\section{Methodology}

\subsection{Dataset}

We use the MD-Offense dataset \citep{leonardelli_agreeing_2021}, designed for the analysis of annotator disagreement in offensive language detection.  It contains 10,753 instances, English tweets annotated by five U.S. annotators per instance.
The data spans three socially and politically salient domains: COVID-19, the 2020 U.S. presidential election, and racial justice, providing rich variability in both content and labeling agreement.
Each instance is assigned a binary label $ c = \{ \texttt{offensive}, \texttt{non-offensive} \} $, obtained via majority vote from five annotators, and denoted thereafter ground-truth label. In addition, each instance is labeled with an agreement level: A++ (5/5), A+ (4/5), or A0 (3/5), which serves as a proxy for linguistic ambiguity \citep{pavlick_inherent_2019, mostafazadeh_davani_dealing_2022}. A++ examples are considered unambiguous, while A0 reflects high annotator disagreement and thus greater ambiguity. The dataset is split into training, validation, and test sets, stratified by domain and agreement level, with full distributions provided in Appendix~\ref{sec:DataAppendix}.
Prior studies report a sharp decline in classification performance as agreement decreases, with F1 scores dropping from approximately 0.91 for A++ to 0.62 for A0 \citep{leonardelli_agreeing_2021}. We replicate this trend on several models, reported in Appendix~\ref{sec:ModelPerformanceAppendix}, underscoring the difficulty of modeling highly ambiguous inputs.

\subsection{Model and Embedding Extraction}

We focus our analysis on RoBERTa-Large, examining both its pretrained and fine-tuned representations. While several encoder-only architectures were fine-tuned as part of our broader experimentation, we focus our analysis on RoBERTa-Large due to strong performance across agreement levels as shown in Appendix~\ref{sec:ModelPerformanceAppendix}  and because focusing on a single model provides clarity of analysis. This makes it a suitable choice for showcasing Mapper's interpretability rather than conducting model benchmarking.
Fine-tuning was performed on the full training set using the ground-truth labels, including all agreement levels from A0 to A++ to reflect the complete distribution of annotator disagreement. While prior work has shown that restricting fine-tuning to A+ and/or A++ subsets can yield higher classification accuracy \citep{leonardelli_agreeing_2021}, our goal is not to optimize predictive performance but to analyze how models structure data in representation space. Fine-tuning was performed using standard hyperparameters: 3 training epochs, a batch size of 8, and a learning rate of $2 \cdot 10^{-5}$. To capture model-internal representations, we extract the \texttt{[CLS]} embedding from the final hidden layer for each test instance because it serves as input to the classifier and encodes task-relevant information.

\subsection{Mapper Parameters}

We implemented the Mapper algorithm using the \texttt{KeplerMapper} Python library \citep{veen_kepler_2019, KeplerMapper_v2.0.1-Zenodo}; 
This library facilitated the projection, covering, and construction of Mapper graphs. To analyze and visualize the resulting networks, we used the \texttt{NetworkX} library \citep{hagberg_exploring_2008}
, which enabled efficient computation of graph-theoretic and topological metrics.
\paragraph{Lens Function}

We chose six lens functions that project high-dimensional instance embeddings of the \texttt{[CLS]}, i.e. \( d = 1024\) into 1D intervals (\( k = 1 \)). Each lens captures a distinct geometric or semantic property of the embedding space.  The centroid projection captures alignment with the decision boundary between offensive and non-offensive classes. PCA highlights global variance, while eccentricity defined as the maximum cosine distance to training points detects peripheral or outlier instances. The L2 norm reflects activation magnitude, potentially linked to confidence. Two random projections serve as baselines to distinguish signal from noise. All lens statistics are computed using the training set only to avoid data leakage. Full definitions are provided in Appendix~\ref{appendix:lens-definitions}.
The use of 1D lenses simplifies comparisons across ambiguity levels by ensuring consistency in Mapper parameters. All graphs are constructed using the same number of intervals, overlap, clustering algorithm, and layout. While higher-dimensional or task-specific lenses may uncover richer structures, we defer such extensions to future work.
While our lenses are geometric, Mapper's topological structure is not derived from the projections themselves. The lens simply partitions the data into overlapping regions, clustering is then performed in the original high-dimensional space. As a result, the Mapper graph reflects how data is connected, not merely how it appears in projection. This shifts the focus from geometric simplification to topological structure, preserving branching, overlaps, and ambiguity patterns often flattened by projection-based methods.

\paragraph{Covering}

We use a fixed resolution of $r = 40$  with 30\% overlap ( \( \epsilon = 0.3 \))  for all lenses. These hyperparameters were selected  after inspection of Mapper graphs across lenses and ambiguity levels. Increasing the number of intervals (e.g., $r > 50$) led to overly fragmented graphs with many disconnected or singleton nodes, while coarse coverings (e.g., $r < 30$) oversimplified the structure, obscuring fine-grained connectivity. A resolution of $r = 40$ offered a middle ground between interpretability and granularity to preserve neighborhood structure without excessive redundancy. The impact of covering resolution on graph topology is provided in Appendix~\ref{appendix:covering-resolution}.
Similarly, for overlap, we tested values in the range $\epsilon \in [0.2, 0.9]$. We avoided higher overlaps as they introduced visual artifacts such as redundant nodes due to duplicate points appearing in adjacent regions, which distorted the underlying topological signal. At the same time, too little overlap caused artificial fragmentation. The choice of $\epsilon = 0.3$ consistently preserved connectivity without inflating the graph. Appendix~\ref{appendix:covering-overlap-hyperparam} provides evidence of how overlap settings influence graph structure.

\paragraph{Clustering} We use HDBSCAN \citep{mcinnes_hdbscan_2017} as our clustering algorithm to avoid imposing a fixed number of clusters. Unlike methods such as \(k\)-means or agglomerative clustering that require a global \(k\), HDBSCAN adapts to the local density of points in each interval, making it well-suited for capturing the heterogeneous structures that Mapper is designed to surface. Fixed-\(k\) clustering often resulted in artificial splits within homogeneous regions, leading to unstable or redundant nodes. In contrast, HDBSCAN dynamically adjusts to local structure and identifies outliers as noise, improving robustness and interpretability. We use cosine distance, appropriate for normalized embedding spaces, and set the minimum cluster size to 2, enabling the detection of fine-grained patterns without excessive fragmentation. Compared to DBSCAN, HDBSCAN requires less manual tuning and generalizes better to varying densities across intervals. The proportion of excluded instances remained moderate and did not meaningfully distort the structure for most lenses, as shown in Appendix~\ref{appendix:hdbscan-noise}, which reports detailed noise rates across lenses and ambiguity levels.

\subsection{Topological Metrics}
We denote the resulting Mapper graph as \( G = (N, E) \), where \( N \) represents the nodes and \( E \) the edges between them. Each node \( n_i \in N \) represents a cluster containing a subset \( X_i = \{x^i_j\} \) of the test instance embeddings. The size of each node is denoted \( S_i = |X_i| \). A connected component is a set of nodes \( C = \{n_i\} \) such that all nodes in \( C \) are connected via edges in \( E \). The full set of data points in a component is denoted \( X_C = \bigcup_{n_i \in C} X_i \), with total size \( |X_C| = \sum_{n_i \in C} S_i \). Based on this formal structure, we propose the following topological evaluation metrics.
\\
\textbf{Component Purity (CP):} average purity of nodes within a connected component \( C \):
\begin{equation}
    \text{CP} = \frac{1}{|X_C|} \sum_{n_i \in C} \sum_{x^i_j \in X_i} \bigl[ y^i_j = c \bigr],
    \label{eq:component-purity}
\end{equation}
where \( y^i_j \) is the label of sample \( x^i_j \), \( c \) is the dominant label (either ground-truth or predicted), and \([P]\) is the Iverson bracket, equal to 1 if \( P \) is true and 0 otherwise. CP measures the extent to which a component is dominated by a single class, accounting for the varying sizes of its nodes.
\\
\textbf{Edge Agreement (EA):} proportion of edges whose endpoints share the same majority class label:
\begin{equation}
    \text{EA} = \frac{1}{|E|} \sum_{(n_i, n_j) \in E} \bigl[ \text{maj}(n_i) = \text{maj}(n_j) \bigr],
    \label{eq:edge-agreement}
\end{equation}
where \( \text{maj}(n_i) \) is the majority ground-truth label among all samples assigned to node \( n_i \). EA captures the global consistency of the Mapper graph by checking how often directly connected nodes share the same class.
\\
\textbf{Majority Match (MM):}
compares predicted vs. true majority labels at the component level:
\begin{equation}
    \text{MM} = \frac{1}{|\mathcal{C}|} \sum_{C \in \mathcal{C}} \bigl[ \text{maj}_{\text{true}}(C) = \text{maj}_{\text{pred}}(C) \bigr],
    \label{eq:majority-match}
\end{equation}
with \( \mathcal{C} \)  the set of connected components in \( G \), and \( \text{maj}_{\text{true}}(C) \), \( \text{maj}_{\text{pred}}(C) \) denote the ground-truth and predicted majority labels in \( X_C \), respectively. 
MM assesses how well the model's predictions align with human consensus at a regional level.

\section{Results and Discussion}

\paragraph{Topological Effects of Fine-Tuning.}

\begin{figure*}[!t]
  \centering
  \begin{subfigure}[t]{0.4\linewidth}
    \centering
    \includegraphics[width=\linewidth]{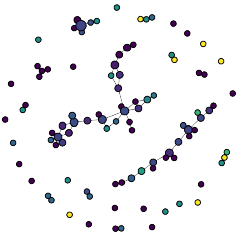}
    \caption{Before fine-tuning.}
    \label{fig:mapper-pca-before-afterA}
  \end{subfigure} \hfill
  \begin{subfigure}[t]{0.4\linewidth}
    \centering
    \includegraphics[width=\linewidth]{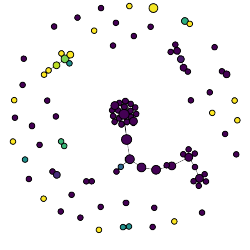}
    \caption{After fine-tuning on the training set.}
    \label{fig:mapper-pca-before-afterB}
  \end{subfigure}
  
  \caption{%
    \label{fig:mapper-pca-before-after}
    Mapper graphs of RoBERTa-Large [CLS] embeddings on the A++ test subset, colored by ground-truth labels (\textcolor[RGB]{72,61,139}{non-offensive}, \textcolor[RGB]{255,215,0}{offensive}, \textcolor[RGB]{32,144,140}{mixed}). Both graphs use a PCA-1D lens (40 intervals, 30\% overlap) and HDBSCAN clustering.
    Nodes are considered \textcolor[RGB]{32,144,140}{mixed} when no single ground-truth label constitutes a majority among their points. 
    Fine-tuning produces a more structured topology: label purity increases, fragmented components are reduced, and class-aligned subregions emerge, suggesting a modular decision space rather than a single global boundary.  
    Note that visual proximity between nodes does not imply embedding similarity, only edges reflect topological closeness through shared overlap.}
\end{figure*}

Fine-tuning restructures the embedding space not by collapsing classes into simple, linearly separable clusters, but by organizing them into modular, non-convex regions. As shown in Figure~\ref{fig:mapper-pca-before-afterA}, the base model exhibits a fragmented topology on the A++ subset, with diffuse clusters and inconsistent labels. Fine-tuning, as shown in Figure~\ref{fig:mapper-pca-before-afterB}, consolidates these components into more coherent regions that are better aligned with class boundaries, yet they remain distributed across disconnected areas.

\begin{figure}[H]
  \centering
  \includegraphics[width=\columnwidth]{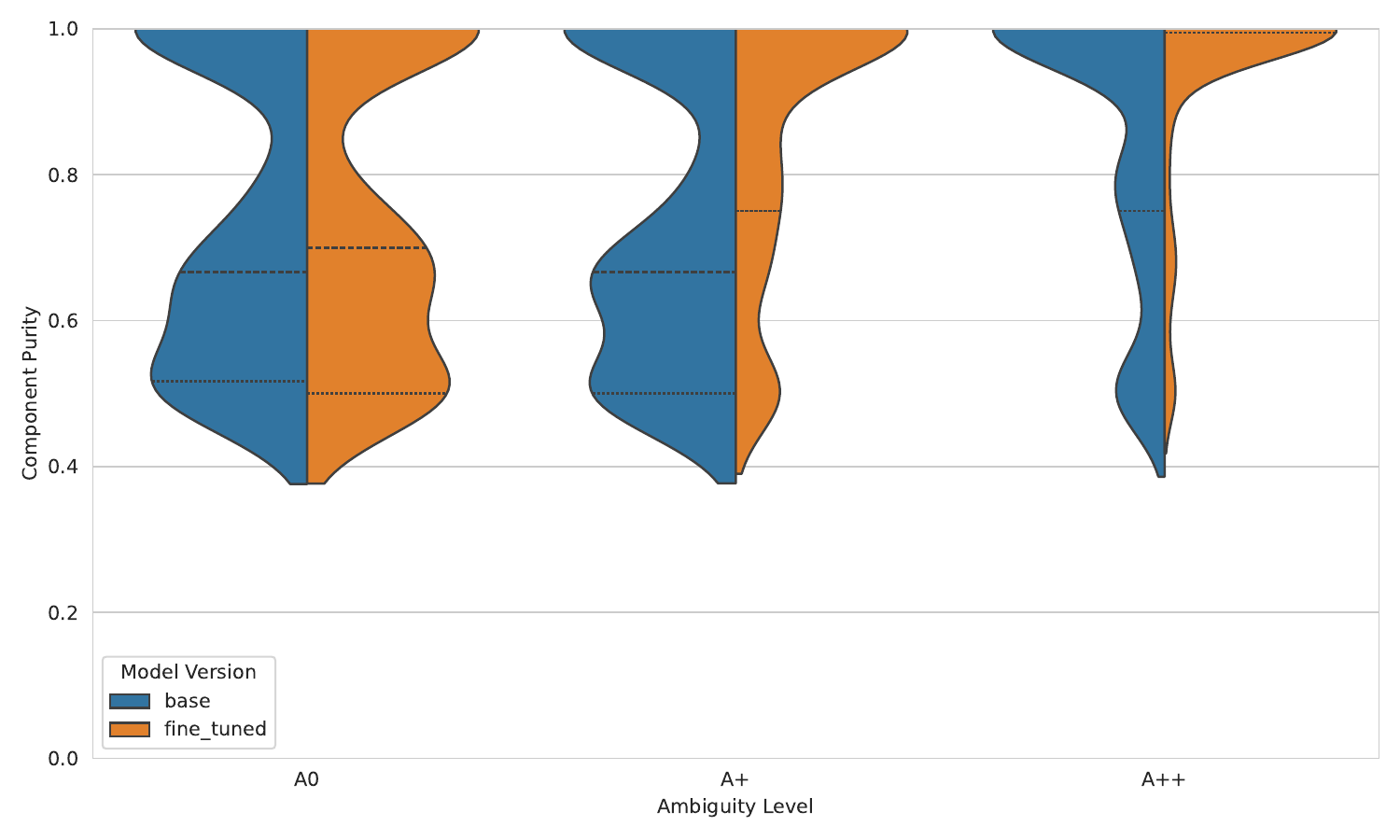}
  \caption{  \label{fig:component_purity_violin}
  \textbf{CP across agreement levels (\textcolor[RGB]{0,0,255}{Base} vs \textcolor[RGB]{255,127,0}{Fine-Tuned}).}
  Distribution of component-level purity scores using eq.~\eqref{eq:component-purity} for each agreement test subset (A0, A+, A++), aggregated across all six 1D lenses. Each point represents the purity of a connected component in a Mapper graph.
  Fine-tuned representations consistently increase class homogeneity within components, especially in A+ and A++, where label agreement is stronger.
  Full per-lens distributions are provided in Appendix~\ref{appendix:per-lens-violin-plots}.
  }
\end{figure}

\begin{figure}[H]
  \includegraphics[width=\columnwidth]{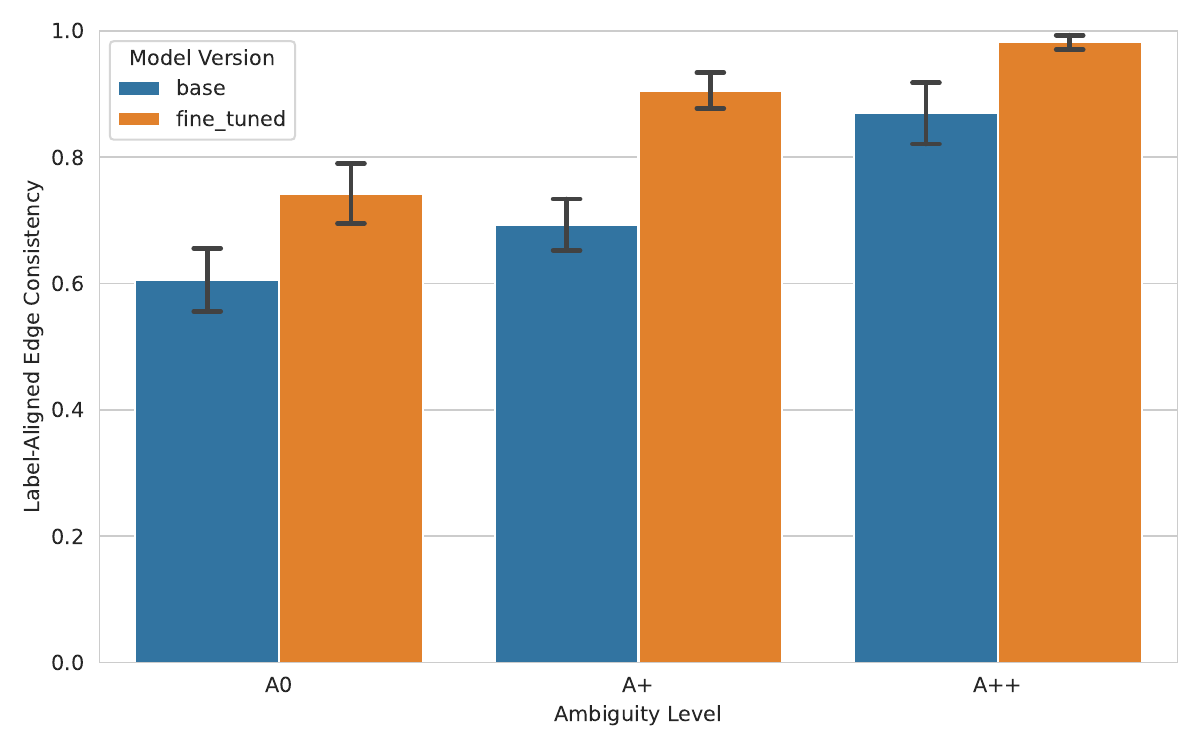}
  \caption{
  \textbf{EA across ambiguity levels (\textcolor[RGB]{0,0,255}{Base} vs \textcolor[RGB]{255,127,0}{Fine-Tuned}).}
  Bar plots show the average EA using eq. ~\eqref{eq:edge-agreement}, computed per Mapper graph and aggregated across all 1D lenses for each greement test subset (A0, A+, A++).
  Fine-tuned embeddings exhibit consistently higher agreement, especially for less ambiguous subsets (A+ and A++), indicating smoother class-consistent transitions between neighboring clusters.
 Error bars denote standard deviation across lenses. Per-lens edge agreement results are reported in Appendix~\ref{appendix:per-lens-edge}, showing that the overall trend holds consistently across all lenses.
  }
  \label{fig:edge_agreement_bar}
\end{figure}

This modularity persists across ambiguity levels. Both component purity and edge agreement increase after fine-tuning in Figures~\ref{fig:component_purity_violin} and  \ref{fig:edge_agreement_bar}, indicating a twofold structural gain: locally, components become more class-homogeneous, and globally, neighboring regions exhibit stronger label consistency. These improvements suggest that the model learns to organize decision boundaries not only within clusters but also across their connections. Gains are weakest in the A0 subset, where annotator disagreement limits geometric consolidation. 

Mapper thus reveals that fine-tuning encodes task-relevant structure without enforcing global convexity. The resulting geometry remains multimodal: the model partitions each class into fragmented, task-aligned subregions. This helps explain why linear probes often fail \citep{hewitt2019designinginterpretingprobescontrol, voita-titov-2020-information}: although widely used, they assume unimodal linear separability and therefore miss the nonlinear, multimodal structure that persists even after training. This highlights the value of topological diagnostics such as Mapper, which are designed to capture the complex organization of decision regions.

\begin{figure*}[ht]
  \centering
  \begin{subfigure}[t]{0.4\linewidth}
    \centering
    \includegraphics[width=\linewidth]{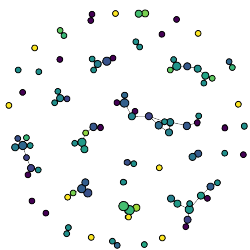}
     \caption{ \label{fig:mapper-a0-gt}
     Ground-truth labels.}
  \end{subfigure} \hfill
  \begin{subfigure}[t]{0.4\linewidth}
    \centering
    \includegraphics[width=\linewidth]{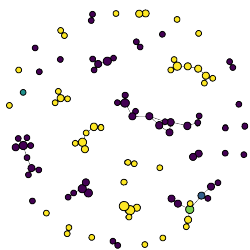}
  \caption{ \label{fig:mapper-a0-mp}
  Model prediction labels.}
  \end{subfigure}
 \caption{%
 \label{fig:mapper-a0-prediction}
  Mapper graphs of [CLS] embeddings on the A0 test subset (RoBERTa-Large fine-tuned on the training set, PCA-1D lens, $r=40$, $\epsilon = 0.3$, HDBSCAN). Left: nodes colored by ground-truth labels, right: same graph colored by model prediction labels,
  (\textcolor[RGB]{68,1,84}{non-offensive}, \textcolor[RGB]{253,231,36}{offensive}, \textcolor[RGB]{32,144,140}{mixed}).  Nodes are considered \textcolor[RGB]{32,144,140}{mixed} when no single ground-truth label constitutes a majority among their points. 
  Although ground-truth labels appear fragmented (left), the model's predicted labels (right) exhibit smoother, regionally consistent patterns reflecting the model's binary classification training objective. Each topological component tends to predict a consistent class, with fewer intra-cluster flips. This suggests that the model performs prediction at the level of coarse topological regions, even in ambiguous data.}
\end{figure*}
\paragraph{Prediction Aligns with Topological Regions.}
In the most ambiguous setting (A0), where label structure is noisiest, fine-tuned models reveal a striking behavior: prediction regions become geometrically coherent. The base model as shown in Figure~\ref{fig:mapper-a0-gt} produces noisy, fragmented label patterns of embeddings. By contrast, the fine-tuned model as shown in Figure~\ref{fig:mapper-a0-mp} imposes smooth region-level decisions: most connected components yield a single predicted label, suggesting classification operates at the level of regions, not individual points. Only one region, in the bottom right of Figure~\ref{fig:mapper-a0-mp}, shows a gradual transition in predictions: it moves from non-offensive to offensive through a small area with mixed labels. This is a rare example of a visible decision boundary in the embedding space. Most other regions are sharply separated, which suggests that the model makes confident, all or nothing predictions at the level of entire regions, even when the data is ambiguous.\\
Table~\ref{tab:component-summary} quantifies this trend: after fine-tuning, over 98\% of components are prediction-pure, even in A0 subset. Despite annotator disagreement, the model confidently projects a single label per region. Yet predictions are not arbitrary,
the majority match rate MM remains high (77.7\% in A0, above 94\% in A+/A++), showing that component-level predictions still align with dominant ground-truth labels. This behavior highlights  two effects. First, the model appears to projects test instances into coarse topological regions that reflect its binary fine-tuning objective. Second,  while this reflects structural confidence, it also explains the model's overconfidence in ambiguous regions: predictions appear to be driven by dominant signals within a component, even when the underlying labels are mixed or uncertain. In A0 test subset, the model confidently assigns a single prediction to all components.This region-level coherence is not visible in scalar metrics such as accuracy or F1 score, but emerges clearly through Mapper's topological lens. Appendix~\ref{appendix:low-purity-components} presents qualitative examples of low-purity components. More broadly, Mapper provides a diagnostic tool to investigate how models group sentences. For instance, whether such groupings reflect meaningful features or unintended artifacts. While we do not pursue this question here, our results highlight it as a promising direction for future work.
\paragraph{Robustness Across Lenses and Models.}
To ensure our findings are not tied to a particular setup, we conducted two robustness checks. 
First, applying 2D lenses produced Mapper graphs and purity trends indistinguishable from the 1D case, as detailed in Appendix~\ref{appendix:2dlens}. 
Second, repeating the analysis with other encoder-only models yielded the same topological patterns observed with RoBERTa-Large: 
modular prediction aligned regions, near-perfect prediction purity, and the largest prediction label gap in A0 as shown in Appendix~\ref{appendix:model2d}. 
Together, these checks indicate that our observations are not artifacts of a particular projection or model, but reflect consistent topological behaviors.

\begin{table*}[htb]
  \centering
  \begin{tabular}{lccc}
    \hline
    \textbf{Ambiguity} 
    & \textbf{CP (GT) (\%)} 
    & \textbf{CP (Pred) (\%)} 
    & \textbf{MM (\%)} \\
    \hline
      A0  & 31.5 $\pm$ 10.1 & 98.0 $\pm$ 2.4   & 77.7 $\pm$ 20.5 \\
    A+  & 54.8 $\pm$ 15.1 & 98.0 $\pm$ 3.2   & 94.1 $\pm$ 12.0 \\
    A++ & 81.8 $\pm$ 10.6 & 99.5 $\pm$ 1.2   & 97.1 $\pm$ 3.9  \\
 
    \hline
  \end{tabular}
  \caption{\label{tab:component-summary}
    Percentage of connected components that exceed the 90\% purity threshold, computed using ground-truth labels (CP~(GT)) and model predictions labels (CP~(Pred)), i.e using eq \eqref{eq:component-purity} with corresponding labels. The column MM reports the proportion of components where the predicted majority label matches the ground-truth majority label (see eq.~\eqref{eq:majority-match}). Results are averaged across all 1D lenses and reported as mean~$\pm$~std. Full per-lens results and additional details about variation are provided in Appendix~\ref{appendix:majority-match-lens}.
}
\end{table*}
\section{Conclusion}

Mapper reveals that fine‑tuning not only compacts and cleans the topology of embedding space, boosting component purity and edge agreement, but also uncovers a modular, non‑convex regions that align closely with model predictions. This reorganization is most visible in unambiguous data (A++), where high agreement among annotators leads to clearly defined, coherent regions. In contrast, ambiguous cases (A0) remain fragmented and structurally noisy, yet are often grouped into confident, homogeneous prediction regions. This suggests that models do not merely reflect label noise, they impose their own structured certainty, even in regions where human disagreement dominates. These findings defy common assumptions about model failure in ambiguity. Rather than randomly guessing, models exhibit systematic region-level behavior, suggesting that ambiguity is encoded and resolved in consistent, if not always accurate ways. Whether this resolution reflects true generalization or an overfit to misleading surface patterns remains an open question.
Topological analysis thus complements scalar metrics by exposing how models internalize uncertainty and reshape representation space. It highlights not only where models succeed, but how they choose to generalize, whether by collapsing disagreement into a dominant class, or by carving complex substructures to navigate ambiguous content. 
Mapper is a promising foundation for developing more reliable and interpretable NLP systems. 
Beyond post-hoc interpretation, these structures could inform proactive modeling strategies, guiding subset selection, fine-tuning regimes, or robustness evaluations focused on ambiguous regions.
By linking embedding geometry, prediction consistency, and human disagreement, Mapper enables a more principled understanding of how models internalize and act on language representations. Leveraging these insights to inform both model development and annotation practices remains a rich direction for future work.

\clearpage  
\section*{Limitations}
While our topological analysis  provides novel insights into how fine-tuned models encode class structure and ambiguity, several limitations should be acknowledged.
\paragraph{Mapper Sensitivity and Interpretability.} 
The Mapper algorithm is highly sensitive to its hyperparameters, particularly the lens function, cover granularity, and clustering strategy. Although we use principled heuristics and conduct lens-wise robustness checks, the interpretability of resulting graphs still depends on these design choices. Topological patterns should therefore be interpreted in relative terms (e.g., trends across training conditions), not as fixed geometric ground truths.

\paragraph{Task and Dataset Specificity.}
Our experiments focus on a single dataset and task (offensive language classification on MD-Offense). 
While this setup allows controlled analysis, it limits generalizability. 
It remains an open question whether similar topological phenomena, such as disconnected decision regions or regional predictive consistency, 
would emerge in other tasks (e.g., natural language inference, sentiment analysis) or with datasets exhibiting different forms of ambiguity.

\paragraph{Lens Expressiveness and Task-Specificity.}
Our analysis primarily relied on simple, interpretable lenses. 
While we extended the study to 2D projections , 
our focus was on general purpose constructions such as PCA, centroid distances, and eccentricity. 
Task specific or adaptive lenses for example, saliency guided projections, uncertainty aware mappings, 
or semantics driven nonlinear transformations may expose different structures, particularly in regions where ambiguity arises from subtle pragmatic or contextual cues. 
Identifying lens designs tailored to the sources of ambiguity in language remains an open direction for future work.

\paragraph{Metric Limitations.}
The structural metrics we propose, component purity, edge agreement, and majority match, capture interpretable geometric trends, but are Mapper-specific and do not directly reflect classification accuracy or generalization. Moreover, they are inherently correlational: improvements in structure after fine-tuning may reflect better task learning, but could also result from overfitting or feature collapse.

\paragraph{Toward Generalizable Topological NLP.}
These limitations are not unique to our study but reflect broader challenges in adapting TDA to NLP. They point to the need for hybrid pipelines combining topological summaries with statistical validation, comparisons to alternative graph structures, and multi-model, multi-task replication.



\bibliography{acl_latex}

\onecolumn
\appendix
\section*{Appendix}
\section{Data Representation}
\label{sec:DataAppendix}
Table~\ref{tab:appendix-data-ratios} summarizes the number of samples and the proportion of offensive examples across dataset splits (train, validation, test) and agreement levels (A0, A+, A++) based on the ground-truth labels. Offensive content is more frequent in the A0 subsets, which may help explain the higher levels of annotator disagreement observed in these cases.
\begin{table*}[!h]
  \centering
  \begin{tabular}{lccc}
    \hline
    \textbf{Split} & \textbf{Agreement Level} & \textbf{Total Samples} & \textbf{Offensive Ratio (\%)} \\
    \hline
    Train & A0  & 1884 & 45.2 \\
          & A+  & 1930 & 32.9 \\
          & A++ & 2778 & 17.1 \\
    \hline
    Validation   & A0  & 322  & 46.9 \\
          & A+  & 317  & 37.2 \\
          & A++ & 465  & 25.6 \\
    \hline
    Test  & A0  & 856  & 45.1 \\
          & A+  & 909  & 39.6 \\
          & A++ & 1292 & 21.1 \\
    \hline
  \end{tabular}
  \caption{\label{tab:appendix-data-ratios}
    Distribution of offensive labels across MD-Offense, grouped by split and agreement level.}
\end{table*}

\section{Model-Level Summary}
\label{sec:ModelPerformanceAppendix}
Table~\ref{tab:appendix-model-metrics} presents test set performance and embedding structure across ambiguity levels (A0, A+, A++) for all fine-tuned models. Accuracy and Macro F1 reflect standard classification performance, while silhouette score variance and Davies–Bouldin index (lower is better) capture structural coherence in embedding space. All metrics are averaged over three random seeds. Best accuracy values are highlighted in bold. RoBERTa-Large consistently achieves strong results across ambiguity levels, motivating its selection for our topological analysis.
\begin{table*}[!h]
\centering
\begin{tabular}{lcccccc}
\hline
\textbf{Model} & \textbf{Agreement Level} & \textbf{Accuracy} & \textbf{F1 Score} & \textbf{Silhouette (±) ↑} & \textbf{DB Index (±) ↓} \\

\hline
BERT-Base     & A0  & 0.607 & 0.602 & 0.327 ± 0.010 & 5.07 ± 0.39 \\
              & A+  & 0.810 & 0.801 & 0.465 ± 0.008 & 1.54 ± 0.03 \\
              & A++ & 0.934 & 0.904 & 0.782 ± 0.014 & 0.83 ± 0.04 \\
\hline
DeBERTa-Base  & A0  & 0.602 & 0.596 & 0.315 ± 0.002 & 5.58 ± 0.11 \\
              & A+  & 0.819 & 0.812 & 0.478 ± 0.012 & 1.41 ± 0.02 \\
              & A++ & 0.943 & 0.916 & 0.801 ± 0.014 & 0.73 ± 0.04 \\
\hline
DistilBERT    & A0  & 0.606 & 0.598 & 0.329 ± 0.006 & 5.43 ± 0.30 \\
              & A+  & 0.811 & 0.801 & 0.472 ± 0.012 & 1.56 ± 0.03 \\
              & A++ & 0.941 & 0.912 & 0.789 ± 0.007 & 0.84 ± 0.02 \\
\hline
E5-Base       & A0  & 0.612 & 0.608 & 0.431 ± 0.003 & 5.60 ± 0.15 \\
              & A+  & 0.809 & 0.800 & 0.489 ± 0.021 & 1.70 ± 0.11 \\
              & A++ & 0.935 & 0.906 & 0.798 ± 0.027 & 0.92 ± 0.11 \\
\hline
MiniLM        & A0  & 0.600 & 0.596 & 0.312 ± 0.003 & 4.56 ± 0.13 \\
              & A+  & 0.790 & 0.783 & 0.461 ± 0.002 & 1.43 ± 0.03 \\
              & A++ & 0.942 & 0.916 & 0.805 ± 0.009 & 0.65 ± 0.03 \\
\hline
RoBERTa-Large & A0  & \textbf{0.615} & 0.608 & 0.441 ± 0.001 & 4.45 ± 0.36 \\
              & A+  & \textbf{0.822} & 0.814 & 0.502 ± 0.041 & 1.16 ± 0.06 \\
              & A++ & \textbf{0.954} & 0.932 & 0.825 ± 0.045 & 0.51 ± 0.04 \\
\hline

\end{tabular}
\caption{\label{tab:appendix-model-metrics}
Test performance and representation metrics for each fine-tuned model.}
\end{table*}

\newpage

\section{Lens Details}
\label{appendix:lens-definitions}

We define the 1D lens functions \( f(x) : \mathbb{R}^d \to \mathbb{R} \) used to project test embeddings \( x \in \mathbb{R}^d \) into scalar values for Mapper analysis. These lens functions serve as the filtering mechanism that guides Mapper's overlapping cover of the data space. Each function highlights a specific geometric or semantic property of the representation space, such as class separation, density, or position relative to cluster centroids. All lens values are computed using statistics from the training set and rescaled to the \([0, 1]\) interval to ensure consistent binning and comparability across lenses.

We use a diverse set of lens types: supervised projections based on class centroids, unsupervised structure via PCA, local and global density indicators like eccentricity or L2 norm, and random directions that simulate generic perspectives of the space. This variety enables a multifaceted topological analysis, helping ensure that observed Mapper structures are not artifacts of a particular projection choice.
\\
Let \( x \in \mathbb{R}^d \) be thereafter the embedding of a test instance.

\begin{itemize}

    \item \textbf{Centroid Projection (\texttt{centroid\_1d})}  
    Projects \( x \) onto the vector connecting the means of the embeddings (centroids) of the two training classes:
    \[
    f(x) =\frac{x^\top (\mu_1 - \mu_0)}{\|\mu_1 - \mu_0\|},
    \]
    with \( \mu_0 \) and \( \mu_1 \) the centroids, i.e. the mean vectors, of the non-offensive \( (y=0) \) and offensive \( (y=1) \) training embeddings subsets, respectively:
    \[
    \mu_y = \frac{1}{|\mathcal{X}_y|} \sum_{x_i \in \mathcal{X}_y} x_i,
    \]
    where \( \mathcal{X}_y \subset \mathbb{R}^d \) is the subset of training embeddings labeled either \( y=0 \)  or \( y=1 \) . The resulting scalar \( f(x) \) reflects how aligned the embedding of the test instance \( x \) is with the class separation direction.

    \item \textbf{PCA First Component (\texttt{pca\_1d})}  
    Projects \( x \) onto the first principal component \( u_1 \) of the training set:
    \[
    f(x) = u_1^\top x,
    \]
    where \( u_1 \in \mathbb{R}^d \) is the direction of maximal variance estimated by applying PCA on the entire training dataset in the embeddings space. This projection captures the dominant geometric trend in the data and often reflects task-relevant variation.

    \item \textbf{Eccentricity (\texttt{eccentricity\_1d})}  
    Measures how distant \( x \) is from the training set, using maximum cosine distance:
    \[
    f(x) = \max_{x_i \in \mathcal{X}_{\texttt{train}}} \left(1 - \frac{x \cdot x_i}{\|x\| \, \|x_i\|} \right),
    \]
    where \( \mathcal{X}_{\texttt{train}} \) denotes the set of all training embeddings. High values of \( f(x) \) indicate that the test point is geometrically distant (i.e., eccentric) from the training distribution.

       \item \textbf{L2 Norm (\texttt{l2norm\_1d})}  
    Measures the Euclidean norm (magnitude) of \( x \) :
    \[
    f(x) = \|x\|_2 = \sqrt{\sum_{i=1}^d x_i^2}.
    \]
    This lens reflects how far the point lies from the origin in embedding space and may capture differences in representation scale or intensity.
    
    \item \textbf{Random Projections (\texttt{random1\_1d} and \texttt{random2\_1d})}  Projects \( x \) onto random direction
\[
f(x) = r^\top x
\]
where \( r \in \mathbb{R}^d \) is a fixed random unit vector sampled once before projection. This lens selects a random direction in the embedding space and projects each test point along it. Different random seeds yield different perspectives of the geometry.
\end{itemize}

\section{Varying Mapper Covering Resolution}
\label{appendix:covering-resolution}
\begin{figure*}[t]
\centering

\begin{subfigure}[b]{0.30\textwidth}
    \includegraphics[width=\linewidth]{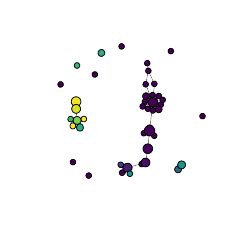}
    \caption{\( r \) = 10}
\end{subfigure}
\hfill
\begin{subfigure}[b]{0.30\textwidth}
    \includegraphics[width=\linewidth]{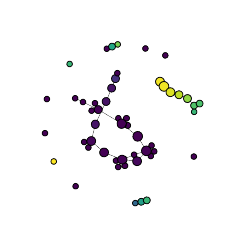}
    \caption{\( r \) = 20}
\end{subfigure}
\hfill
\begin{subfigure}[b]{0.30\textwidth}
    \includegraphics[width=\linewidth]{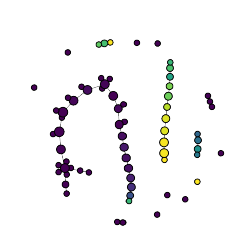}
    \caption{\( r \) = 30}
\end{subfigure}

\vspace{1em}

\begin{subfigure}[b]{0.30\textwidth}
    \includegraphics[width=\linewidth]{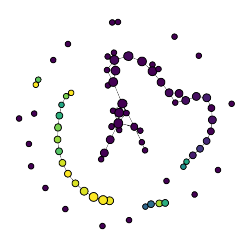}
    \caption{\( r \) = 40}
\end{subfigure}
\hfill
\begin{subfigure}[b]{0.30\textwidth}
    \includegraphics[width=\linewidth]{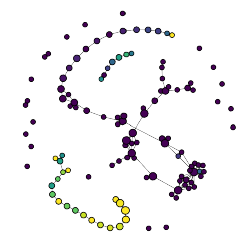}
    \caption{\( r \) = 50}
\end{subfigure}
\hfill
\begin{subfigure}[b]{0.30\textwidth}
    \includegraphics[width=\linewidth]{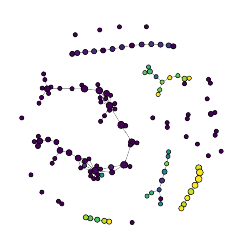}
    \caption{\( r \) = 60}
\end{subfigure}

\vspace{1em}

\begin{subfigure}[b]{0.30\textwidth}
    \includegraphics[width=\linewidth]{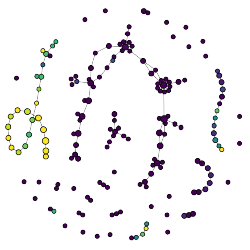}
    \caption{\( r \) = 70}
\end{subfigure}
\hfill
\begin{subfigure}[b]{0.30\textwidth}
    \includegraphics[width=\linewidth]{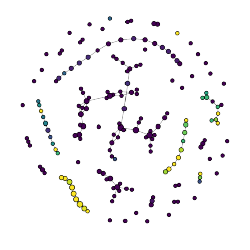}
    \caption{\( r \) = 80}
\end{subfigure}
\hfill
\begin{subfigure}[b]{0.30\textwidth}
    \includegraphics[width=\linewidth]{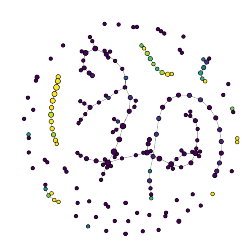}
    \caption{\( r \) = 90}
\end{subfigure}

\caption{\label{fig:covering-resolution}
Effect of increasing the number of intervals \( r \) in the Mapper covering. Higher resolution introduces finer granularity and reveals smaller connected regions, but can also lead to fragmentation and spurious nodes. All graphs use the same lens and fixed overlap (\(\epsilon = 0.3\)).}
\end{figure*}

Figure~\ref{fig:covering-resolution} shows the effect of increasing the number of intervals \( r \) in the Mapper covering on the same dataset subset (A++) using the \texttt{random1\_1d} lens. All graphs were generated with the same HDBSCAN configuration and a fixed overlap of \(\epsilon = 0.3\).
As \( r \) increases from 10 to 90, the representation space becomes more finely partitioned. Low values (e.g., 10–20) result in overly coarse graphs that merge distinct regions and obscure local structure. In contrast, high values (e.g., 80–90) lead to significant fragmentation, with many disconnected or singleton components. This reflects how excessive resolution can distort the underlying topological signal.
A middle setting, such as \( r \) = 30 or 40, consistently offers a good trade-off between local granularity and global connectivity. While this trend holds across models and ambiguity levels, the degree of fragmentation also depends on the lens and the noisiness of the embedding space. These hyperparameter choices are thus best calibrated in relation to both the geometry of the data and the interpretability needs of the analysis.

\section{Varying Mapper Overlap}
\label{appendix:covering-overlap-hyperparam}
\begin{figure*}[t]
\centering

\begin{subfigure}[b]{0.30\textwidth}
    \includegraphics[width=\linewidth]{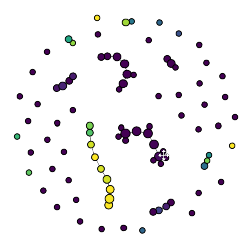}
   \caption{Overlap 10\%}

\end{subfigure}
\hfill
\begin{subfigure}[b]{0.30\textwidth}
    \includegraphics[width=\linewidth]{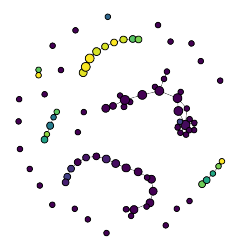}
    \caption{Overlap 20\%}

\end{subfigure}
\hfill
\begin{subfigure}[b]{0.30\textwidth}
    \includegraphics[width=\linewidth]{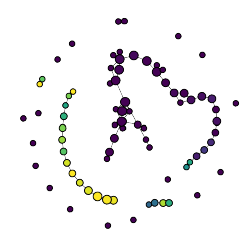}
    \caption{Overlap 30\%}
\end{subfigure}

\vspace{1em}

    \begin{subfigure}[b]{0.30\textwidth}
        \includegraphics[width=\linewidth]{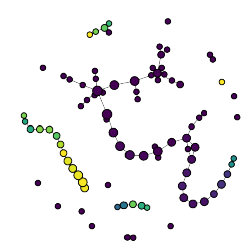}
        \caption{Overlap 40\%}
    \end{subfigure}
\hfill
    \begin{subfigure}[b]{0.30\textwidth}
        \includegraphics[width=\linewidth]{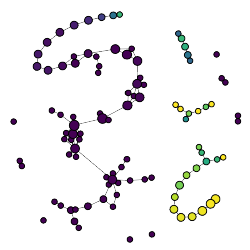}
        \caption{Overlap 50\%}
    \end{subfigure}
\hfill
    \begin{subfigure}[b]{0.30\textwidth}
        \includegraphics[width=\linewidth]{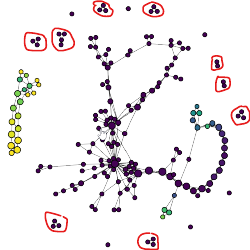}
        \caption{Overlap 60\%}
    \end{subfigure}

\vspace{1em}

    \begin{subfigure}[b]{0.30\textwidth}
        \includegraphics[width=\linewidth]{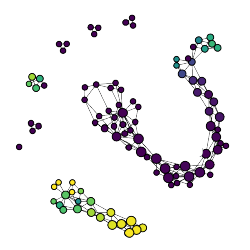}
        \caption{Overlap 70\%}
    
    \end{subfigure}
\hfill
    \begin{subfigure}[b]{0.30\textwidth}
        \includegraphics[width=\linewidth]{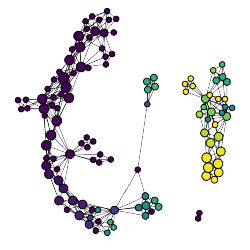}
        \caption{Overlap 80\%}
    
    \end{subfigure}
\hfill
    \begin{subfigure}[b]{0.30\textwidth}
        \includegraphics[width=\linewidth]{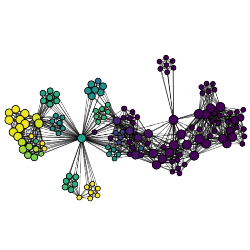}
        \caption{Overlap 90\%}
    \end{subfigure}

\caption{\label{fig:covering-overlap}
Effect of varying Mapper overlap $\epsilon$ from 0.1 to 0.9 on the A++ subset, using the \texttt{random1\_1d} lens and fixed resolution (\( r \) = 40). Low overlap causes fragmentation, while high overlap leads to inflated graphs. Overlap around 30\% provides the best trade-off.}

\end{figure*}

Figure~\ref{fig:covering-overlap} illustrates the effect of increasing Mapper overlap from 10\% to 90\%, using the \texttt{random1\_1d} lens on the A++ subset and a fixed resolution of \( r \) = 40. All graphs were generated with identical HDBSCAN settings to isolate the effect of overlap alone.
At very low values (10\%–20\%), the covering becomes overly rigid, leading to severe fragmentation: adjacent regions are treated as disconnected, resulting in a large number of small or isolated components. As overlap increases to 30\%–40\%, local continuity improves and previously disconnected regions begin to merge into coherent graph structures. This is the range where Mapper begins to meaningfully recover the underlying geometry without inflating the graph.
Beyond 50\%, however, the increasing redundancy from duplicated points across overlapping bins causes a rapid inflation in graph complexity. This leads to spurious edges, tightly packed node clusters, and unnatural fan-like structures, particularly visible at 80\%–90\%. These artifacts distort the topology and reduce interpretability by over-amplifying minor variations.
Overall, the 30\%–40\% overlap range provides the best trade-off: it allows components to remain connected without over-smoothing the embedding space. This setting is therefore used consistently in our Mapper analyses.

\section{HDBSCAN Noise Exclusion by Lens}
\label{appendix:hdbscan-noise}
Table~\ref{tab:appendix-hdbscan-noise} reports the proportion of test points labeled as noise by HDBSCAN across 1D lenses and ambiguity levels. Eccentricity consistently yields the lowest exclusion rates often under 5\%, reflecting its inclusive behavior due to its definition as the maximum cosine distance to the training set. It captures structurally atypical inputs without discarding them. Random projections also result in minimal exclusion, as expected from their agnostic geometric alignment.\\
In contrast, task-oriented lenses such as \texttt{centroid\_1d} and \texttt{pca\_1d} exhibit noticeably higher noise after fine-tuning. These lenses emphasize class-separating directions, and the increase in excluded points likely reflects sharper representational boundaries introduced during supervised optimization. Ambiguous or boundary-adjacent instances are more frequently labeled as noise in these projections.\\
These patterns align with our overall claims: fine-tuning enhances topological regularity but also increases exclusion in class-aligned projections. Across lenses, noise remains within acceptable bounds and does not distort Mapper structure. HDBSCAN thus effectively balances robustness and interpretability in our setup.

\begin{table}[!h]
\centering
\begin{tabular}{lccc|ccc}
\hline
\textbf{Lens} & \multicolumn{3}{c}{\textbf{Base Noise (\%)}} & \multicolumn{3}{c}{\textbf{Fine-tuned Noise (\%)}} \\
&A0&A+&A++&A0&A+&A++\\ 
\hline
\texttt{centroid\_1d} & 20.1\%&  9.5\%&  8.3\% &  26.6\%&  27.4\%&  20.0\% \\
\texttt{pca\_1d}              & 20.7\%&  13.9\%&  9.3\% &  26.0\%&  26.7\%&  20.3\% \\
\texttt{eccentricity\_1d}     & 10.0\%&  8.5\%&  5.7\% &  3.8\%&  2.1\%&  2.1\% \\
\texttt{l2norm\_1d}           & 21.3\%&  12.4\%&  19.1\% &  5.6\%&  5.8\%&  9.4\% \\
\texttt{random1\_1d}          & 2.9\%&  1.7\%&  1.2\% &  7.8\%&  9.2\%&  10.5\% \\
\texttt{random2\_1d}          & 2.2\%&  1.9\%&  2.3\% &  6.3\%&  6.5\%&  7.1\% \\
\hline
\end{tabular}

\caption{\label{tab:appendix-hdbscan-noise}
Proportion of test points labeled as noise by HDBSCAN for each 1D lens, reported inline by ambiguity level (A0, A+, A++). Values correspond to final RoBERTa-Large representations and are not averaged across seeds.}
\end{table}

\section{Violin plots per lens for component purity}
This figure complements Figure~\ref{fig:component_purity_violin} by breaking down purity distributions per lens. Despite variation in Mapper construction, the core trend holds: fine-tuning improves the local class consistency (purity) of topological components across all lens types. This robustness supports the interpretation that observed topological shifts reflect meaningful representational changes rather than artifacts of a specific projection.
\label{appendix:per-lens-violin-plots}
\begin{figure}[h]
  \includegraphics[width=\columnwidth]{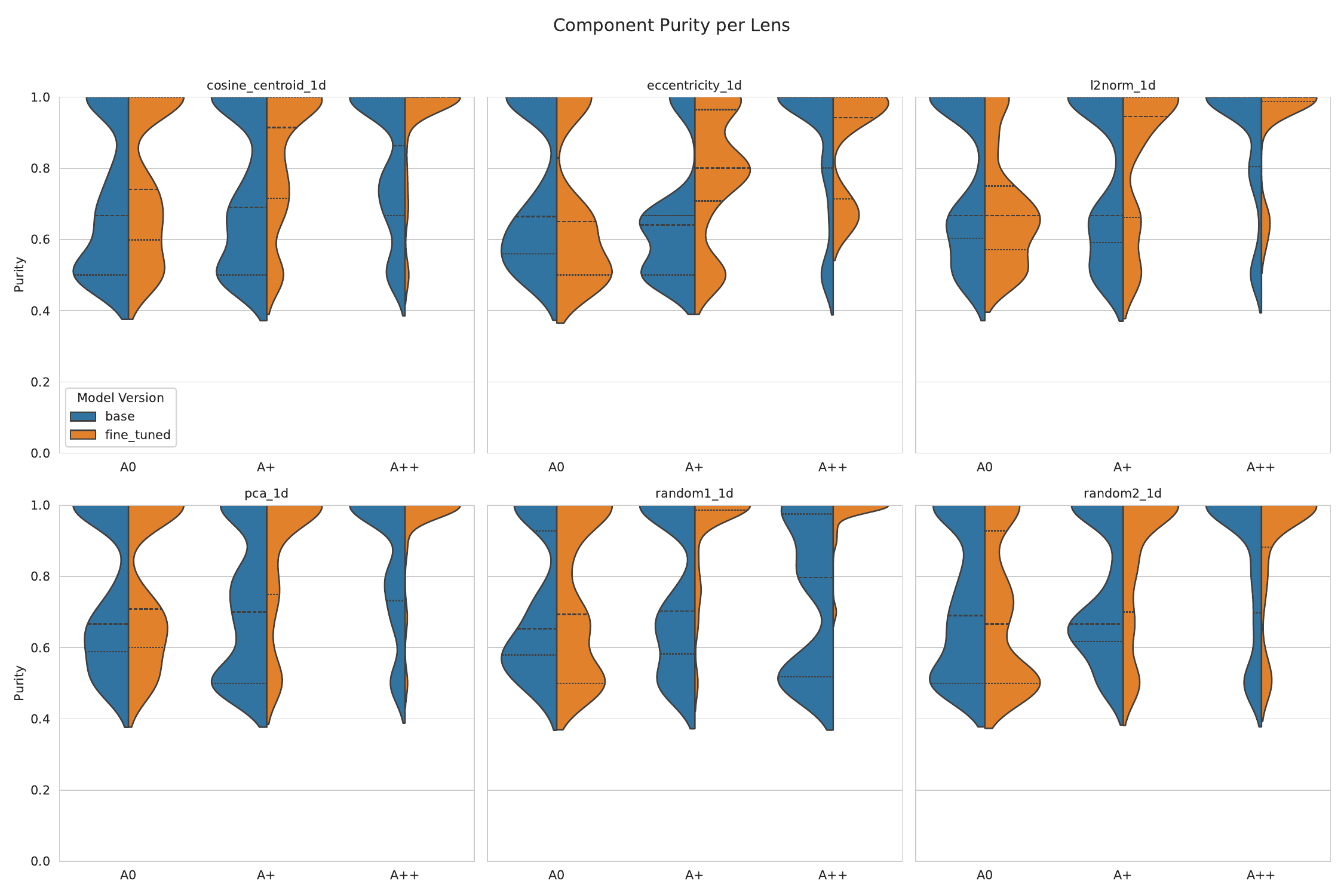}
   \caption{Component purity distributions across 1D lenses. Each subplot shows the purity of connected components for \textcolor[RGB]{0,0,255}{base} vs \textcolor[RGB]{255,127,0}{fine-Tuned} models, separated by ambiguity level (A0, A+, A++). Fine-tuning consistently increases purity across all lenses, particularly for less ambiguous data (A+ and A++). Improvements are visible even for random lenses, suggesting the effect is not lens-specific. However, A0 remains harder to organize topologically, with more dispersed purity scores.}
\end{figure}

\section{Edge Agreement per Lens}
\label{appendix:per-lens-edge}
\begin{figure}[h]
  \includegraphics[width=\columnwidth]{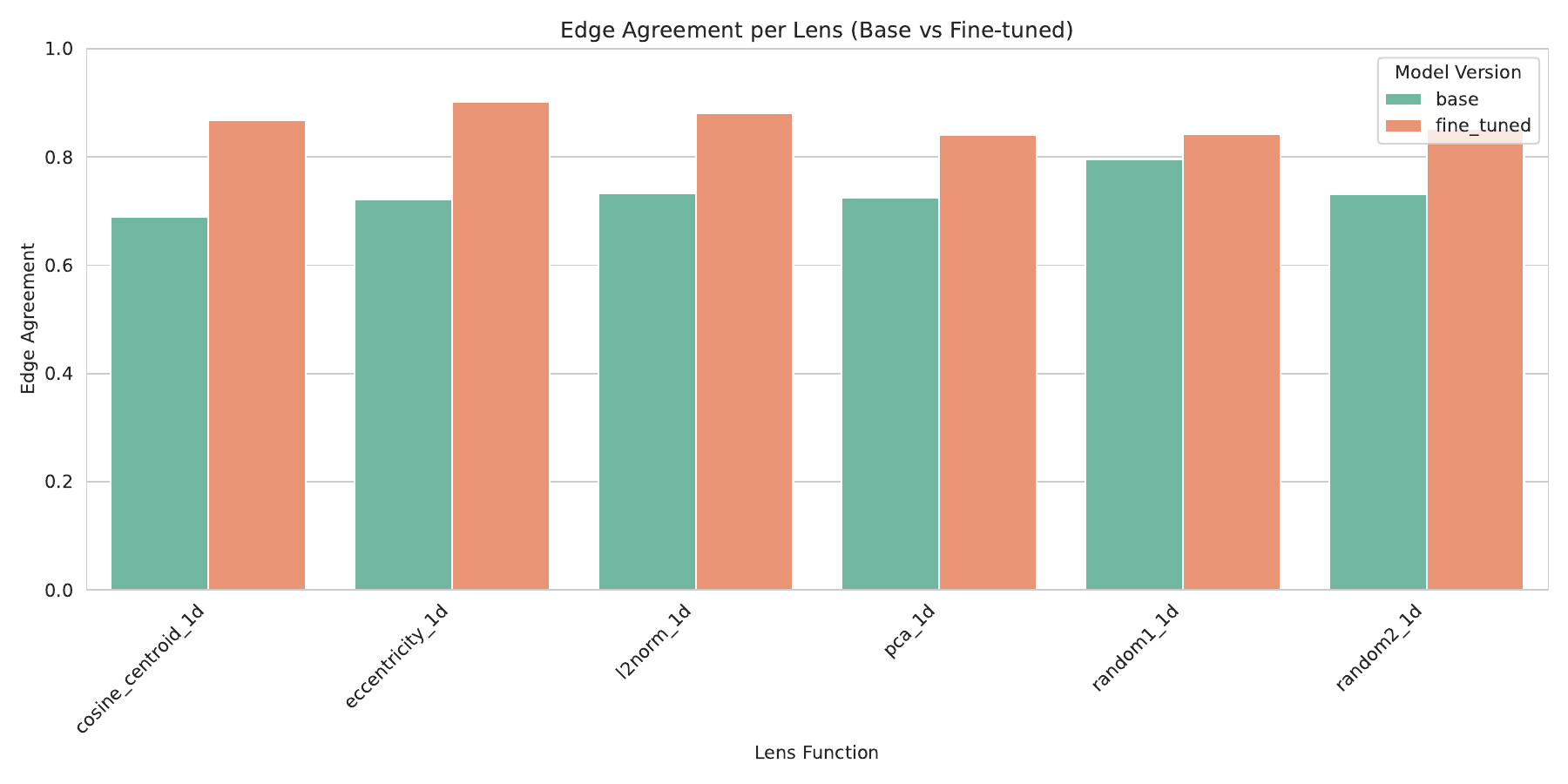}
   \caption{ \label{fig:edge-agreement-lens}
   Edge agreement across lenses for base and fine-tuned RoBERTa-Large models. Each bar reports the average proportion of edges connecting nodes with the same majority class label, across Mapper graphs constructed using different 1D lens functions. Fine-tuned models consistently exhibit higher edge agreement across all lenses, reflecting smoother topological class boundaries.}
\end{figure}

Figure~\ref{fig:edge-agreement-lens} displays edge agreement scores disaggregated by lens type. While the main paper aggregates across lenses to report overall trends, this breakdown reveals how lens choice influences topological structure. All lenses benefit from fine-tuning, but the extent of improvement varies: more structured lenses like \texttt{eccentricity\_1d} and \texttt{l2norm\_1d} show clearer class-consistent transitions than random projections. These differences highlight the importance of lens selection when using Mapper to interpret model geometry.

\section{Per-Lens Analysis of Component Purity and Prediction Alignment}
\label{appendix:majority-match-lens}

Table~\ref{tab:appendix-lens-breakdown} reports component-level metrics broken down by ambiguity level and Mapper lens type. While the main paper focuses on averages aggregated across all lenses (Table~\ref{tab:component-summary}) for robustness, this detailed view helps interpret the observed variance and highlights lens-specific behavior.

Not all lenses yield equally structured Mapper graphs. For example, \texttt{pca\_1d} and \texttt{centroid\_1d} consistently produce high-quality decompositions, with strong purity and high majority match across ambiguity levels. In contrast, lenses like \texttt{l2norm\_1d} or \texttt{random2\_1d} sometimes produce overly coarse or fragmented graphs, especially on ambiguous data (A0). This leads to volatile purity scores and sharp drops in alignment with ground truth. For example, only 50\% majority match for \texttt{l2norm\_1d} in A0.
Interestingly, even lenses like \texttt{eccentricity\_1d}, which exhibit consistently high prediction purity and perfect majority match, can do so while capturing only limited ground-truth variation. These cases underscore that high predictive agreement does not always imply faithful alignment with human annotations, especially in ambiguous regions.

Such differences illustrate why lens choice matters: while Mapper preserves more structure than projection-based methods, the specific lens determines which aspects of the geometry are emphasized. This explains the standard deviations reported in Table~\ref{tab:component-summary} and motivates our decision to report aggregated trends in the main analysis, while offering full per-lens breakdowns here for transparency and interpretability.

\begin{table*}[t]
  \centering
  \begin{tabular}{llccc}
    \hline
    \textbf{Ambiguity} & \textbf{Lens} & \textbf{CP (GT) (\%)}  & \textbf{CP (Pred) (\%)}  & \textbf{MM (\%)} \\
    \hline
     A0  & centroid\_1d & 40.0 & 100.0 & 81.2 \\
        & pca\_1d              & 42.5 & 97.5  & 75.0  \\
        & eccentricity\_1d     & 28.6 & 100.0 & 100.0 \\
        & l2norm\_1d           & 15.4 & 100.0 & 50.0  \\
        & random1\_1d          & 36.4 & 95.5  & 100.0 \\
        & random2\_1d          & 26.3 & 94.7  & 60.0  \\
    \hline
    A+  & centroid\_1d & 50.0 & 100.0 & 100.0 \\
        & pca\_1d              & 60.6 & 100.0 & 100.0 \\
        & eccentricity\_1d     & 33.3 & 93.3  & 100.0 \\
        & l2norm\_1d           & 50.0 & 100.0 & 100.0 \\
        & random1\_1d          & 79.2 & 100.0 & 94.7  \\
        & random2\_1d          & 55.6 & 94.4  & 70.0  \\
    \hline
    A++ & centroid\_1d & 82.9 & 100.0 & 96.6 \\
        & pca\_1d              & 89.8 & 100.0 & 100.0 \\
        & eccentricity\_1d     & 66.7 & 100.0 & 100.0 \\
        & l2norm\_1d           & 83.3 & 100.0 & 100.0 \\
        & random1\_1d          & 95.2 & 100.0 & 90.0 \\
        & random2\_1d          & 72.7 & 97.0  & 95.8 \\
     \hline
   
  \end{tabular}
  \caption{\label{tab:appendix-lens-breakdown}
    Per-lens breakdown of component-level metrics: percentage of components exceeding the 90\% purity threshold for ground-truth and model predictions, and prediction–label majority match among pure components (see eq. ~\eqref{eq:majority-match}).
  }
\end{table*}

\clearpage
\section{Qualitative Examples from Cluster Analysis}
\label{appendix:low-purity-components}
To better understand how models handle ambiguous or low-consensus regions, we examine individual examples from Mapper components with low ground-truth purity (i.e., clusters mixing offensive and non-offensive samples). All predictions are from the fine-tuned RoBERTa-Large model.

These examples show that consistent model predictions often emerge even when annotators disagree. In some cases, this may reflect the model's reliance on strong surface cues, in others, it may suggest that distinct linguistic phenomena (e.g., sarcasm, political critique, racial discourse) are projected into overlapping regions of embedding space.

Mapper allows us to identify and explore such structures. Whether these regions reflect meaningful generalization or model overconfidence remains an open question. By surfacing how instances are grouped and predicted together, Mapper provides a useful entry point for deeper error analysis and for assessing how models internalize disagreement.

\begin{description}
\item[\textbf{Example 1 (Domain 1 - Social Issues/Race)}] \hfill \\
\textbf{Text:} \textit{``Does this black life matter to BLM? [...] BLACK AMERICA, look at what Sharpton is doing to you, YOU MEAN NOTHING TO HIM.''} \\
\textbf{Annotations:} \texttt{1,1,0,1,0} (3 offensive, 2 non-offensive) \\
\textbf{Ground-truth label:} Offensive (1) \\
\textbf{Model prediction label:} Non-offensive (0) with \textbf{97.2\% confidence}

\item[\textbf{Example 2 (Domain 0 - Political Discourse)}] \hfill \\
\textbf{Text:} \textit{``President Trump's rallies [...] Have you not even seen Dementia-Joe??''} \\
\textbf{Annotations:} \texttt{0,0,1,0,1} (2 offensive, 3 non-offensive) \\
\textbf{Ground-truth label:} Non-offensive (0) \\
\textbf{Model prediction label:} Non-offensive (0) with \textbf{97.1\% confidence}

\item[\textbf{Example 3 (Domain 2 - COVID-19/Health)}] \hfill \\
\textbf{Text:} \textit{``We will not survive 2 more years of Gov DeSantis; he's really out here tryna kill people [...]''} \\
\textbf{Annotations:} \texttt{1,1,1,0,0} (3 offensive, 2 non-offensive) \\
\textbf{Ground-truth label:} Offensive (1) \\
\textbf{Model prediction label:} Non-offensive (0) with \textbf{96.4\% confidence}

\end{description}
\section{Robustness with 2D Lenses}
\label{appendix:2dlens}
\begin{figure}[h]
  \includegraphics[width=\columnwidth]{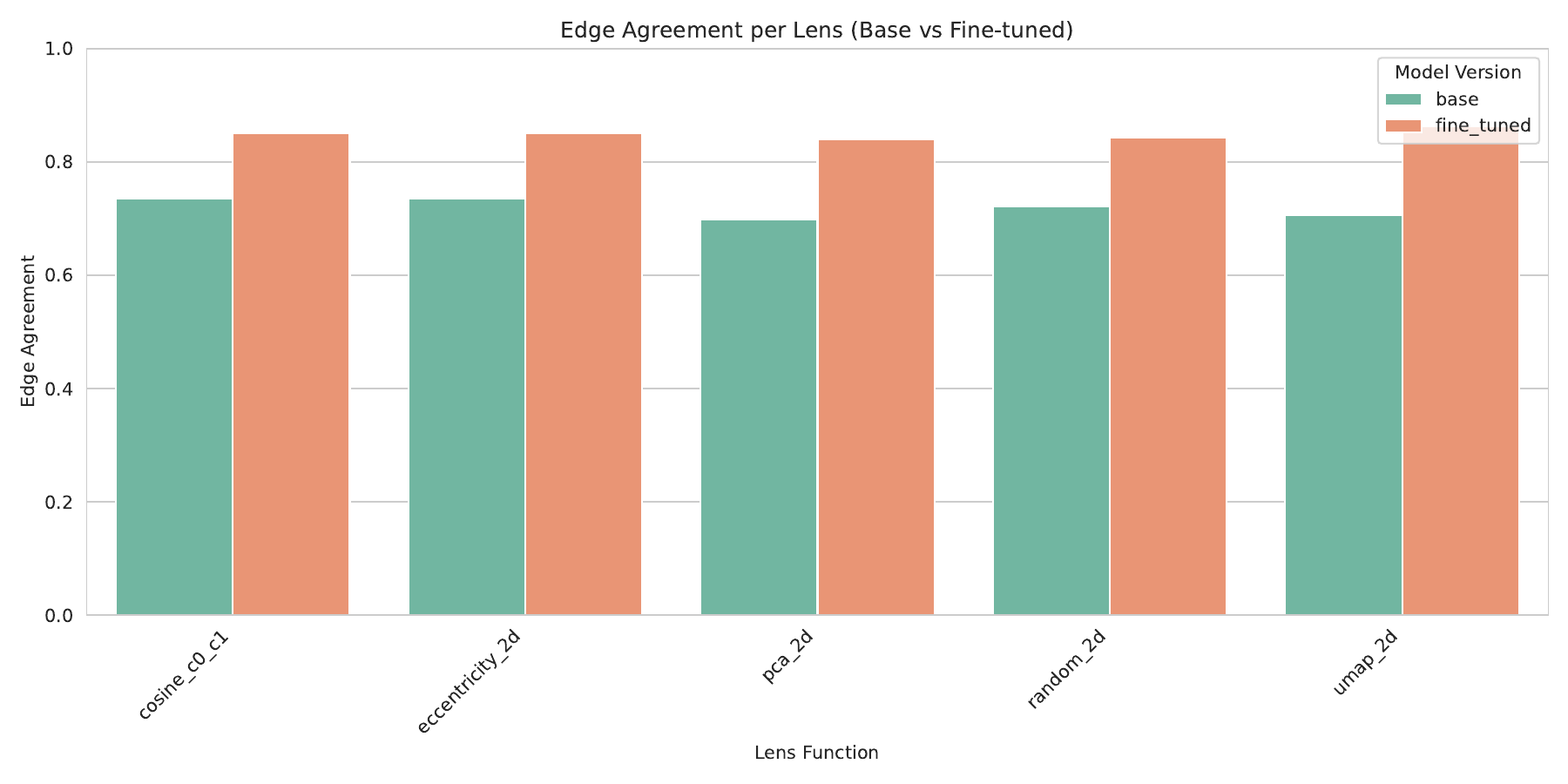}
   \caption{ \label{fig:edge-agreement-lens-2D}
   Edge agreement across 2D lenses for base and fine-tuned RoBERTa-Large models. Fine-tuning consistently increases edge agreement, indicating smoother and more coherent decision boundaries.}
\end{figure}

\begin{figure}[h]
  \includegraphics[width=\columnwidth]{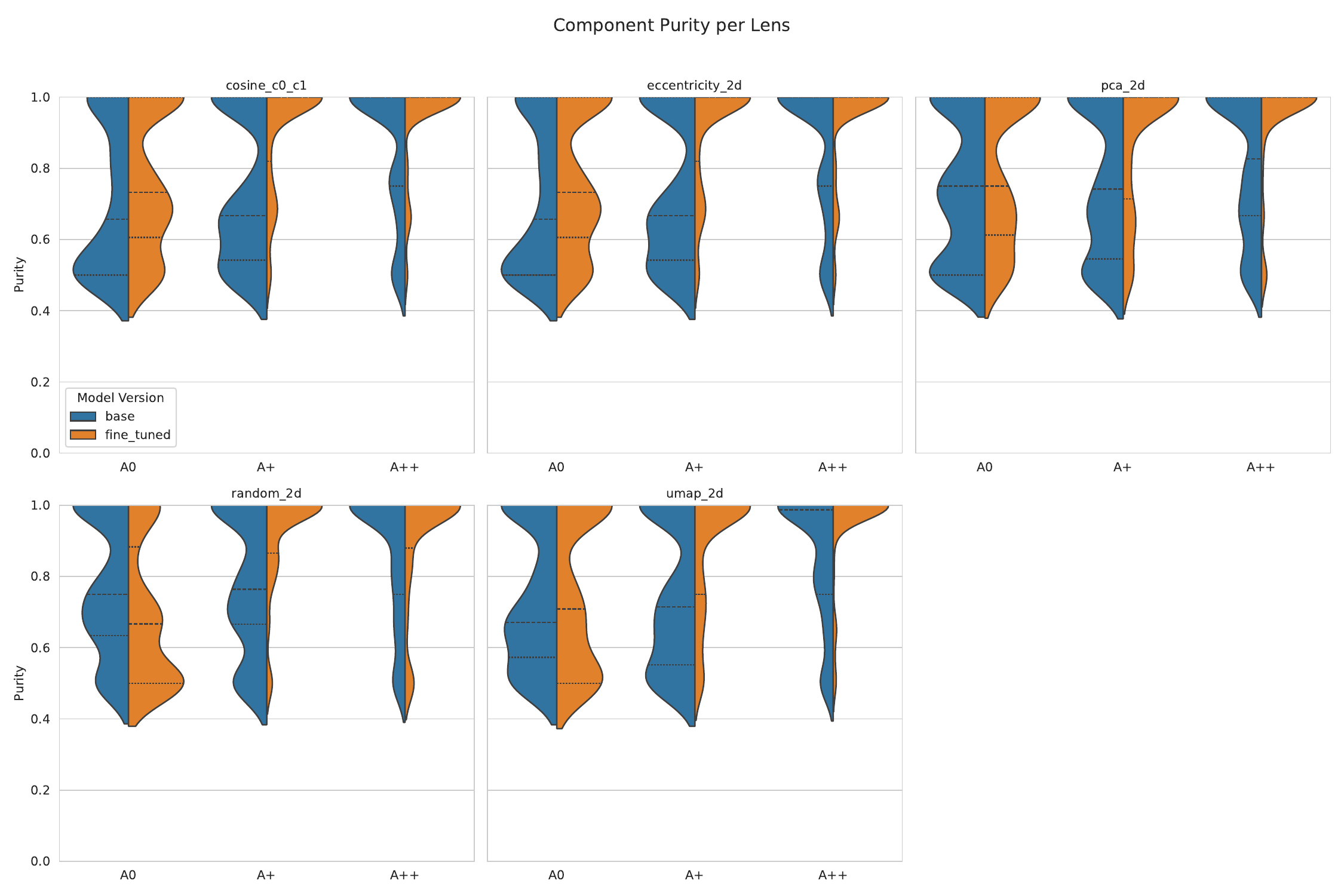}
   \caption{ \label{fig:violin-agreement-lens-2D}
     Distribution of component purity across 2D lenses for RoBERTa-Large.  Consistent with the 1D case, fine-tuned embeddings yield near-perfect prediction purity, while label purity decreases most sharply in A0.}
\end{figure}

While the main study emphasized one-dimensional lenses for their interpretability and exploratory value, here we provide supplementary results using two-dimensional projections. Consistent with the 1D case, fine-tuned embeddings yield near-perfect prediction purity, while label purity decreases most sharply in A0. These trends hold across all five 2D lenses tested (Table~\ref{tab:table_2d}), with component purity distributions and edge agreement metrics (Figures~\ref{fig:violin-agreement-lens-2D}, \ref{fig:edge-agreement-lens-2D}) confirming that, for this dataset, the structural organization revealed by Mapper is stable across projection dimensionality. For 2D lenses, we set Mapper hyperparameters to $r=20$ and $\epsilon=0.2$ to ensure graphs were neither too fragmented nor too connected, following the same consistency checks as in the 1D case. While additional or task-specific lenses (e.g., 3D or uncertainty-aware projections) might expose different structures, our results suggest that the core topological behavior is consistent across 1D and 2D lenses in this setting.
\begin{table*}
  \centering
 \begin{tabular}{llccc}
    \hline
      \textbf{Ambiguity} & \textbf{Lens} & \textbf{CP (GT) (\%)}  & \textbf{CP (Pred) (\%)}  & \textbf{MM (\%)} \\
    \hline
    A0  & cosine\_c0\_c1   & 35.4 & 100.0 & 88.2 \\
        & eccentricity\_2d & 35.4 & 100.0 & 88.2 \\
        & pca\_2d          & 44.2 & 100.0 & 68.4 \\
        & random\_2d       & 25.7 & 100.0 & 77.8 \\
        & umap\_2d         & 39.2 &  95.9 & 77.8 \\
    \hline
    A+  & cosine\_c0\_c1   & 71.9 & 100.0 & 100.0 \\
        & eccentricity\_2d & 71.9 & 100.0 & 100.0 \\
        & pca\_2d          & 63.8 & 100.0 & 94.6 \\
        & random\_2d       & 69.0 & 100.0 & 95.0 \\
        & umap\_2d         & 66.2 &  97.5 & 94.3 \\
    \hline
    A++ & cosine\_c0\_c1   & 81.1 & 100.0 & 100.0 \\
        & eccentricity\_2d & 81.1 & 100.0 & 100.0 \\
        & pca\_2d          & 82.0 & 100.0 & 100.0 \\
        & random\_2d       & 73.5 & 100.0 & 100.0 \\
        & umap\_2d         & 86.3 & 100.0 & 100.0 \\
    \hline
  \end{tabular}

  \caption{\label{tab:table_2d} Percentage of connected components that exceed the 90\% purity threshold, computed using ground-truth labels (CP~(GT)) and model predictions labels (CP~(Pred)), i.e using eq \eqref{eq:component-purity} with corresponding labels. The column MM reports the proportion of components where the predicted majority label matches the ground-truth majority label.}

\end{table*}

\clearpage
\section{Robustness Across Models}
To test robustness across architectures, we repeated the Mapper analysis on additional encoder-only models: BERT-Base, DeBERTa-Base, DistilBERT, E5-Base, and MiniLM, alongside RoBERTa-Large. Table~\ref{tab:models_appendix} reports component purity aggregated across lenses, with values shown as mean ± standard deviation. Across all models, the qualitative pattern is consistent: prediction purity remains higher than label purity, and the largest prediction label gap occurs in A0. This indicates that models tend to impose coherent predictions at the level of topological regions, following dominant signals within components rather than the full variability of annotator labels. While smaller models such as BERT-Base and DistilBERT achieve lower absolute prediction purity, their overall topological organization mirrors that of RoBERTa-Large, confirming that our findings are not model-specific.

\label{appendix:model2d}
\begin{table*}[htb]
  \centering
  \begin{tabular}{llccc}
    \hline
      \textbf{Ambiguity} & \textbf{Lens} & \textbf{CP (GT) (\%)}  & \textbf{CP (Pred) (\%)}  & \textbf{MM (\%)} \\
    \hline
    RoBERTa-Large & A0 & 31.5 ±10.1 & 98.0 ± 2.4   & 77.7 ± 20.5 \\
                  &  A+  & 54.8± 15.1 & 98.0 ± 3.2   & 94.1 ± 12.0 \\
    & A++ & 81.8 ±10.6 & 99.5 ± 1.2   & 97.1± 3.9  \\
                 
    \hline
    BERT-Base     & A0 & 30.2 ± 10.7 & 59.2 ± 19.0 & 67.0 ± 4.9 \\
                  & A+ & 42.8 ± 18.4 & 54.2 ± 17.6 & 77.0 ± 8.2 \\
                  & A++ & 68.1 ± 10.2 & 68.6 ± 7.2 & 86.3 ± 8.8 \\
    \hline
    DeBERTa-Base  & A0 & 31.1 ± 9.1 & 88.4 ± 10.7 & 68.2 ± 9.4 \\
                  & A+ & 52.9 ± 21.7 & 95.1 ± 5.5 & 85.8 ± 9.3 \\
                  & A++ & 72.4 ± 15.5 & 95.0 ± 4.6 & 82.6 ± 8.5 \\
    \hline
    DistilBERT    & A0 & 28.3 ± 12.5 & 62.8 ± 17.9 & 54.8 ± 15.4 \\
                  & A+ & 48.2 ± 8.2 & 65.6 ± 6.9 & 69.1 ± 5.3 \\
                  & A++ & 65.0 ± 15.2 & 66.7 ± 19.7 & 85.0 ± 10.6 \\
    \hline
    E5-Base       & A0 & 31.7 ± 6.2 & 78.3 ± 14.1 & 56.3 ± 6.5 \\
                  & A+ & 48.5 ± 15.0 & 80.8 ± 17.1 & 68.6 ± 12.6 \\
                  & A++ & 69.6 ± 10.2 & 91.1 ± 6.3 & 74.9 ± 11.5 \\
    \hline
    MiniLM        & A0 & 34.5 ± 5.4 & 83.3 ± 13.2 & 61.5 ± 2.6 \\
                  & A+ & 48.6 ± 15.8 & 93.0 ± 5.5 & 77.9 ± 7.1 \\
                  & A++ & 73.3 ± 16.8 & 94.5 ± 4.3 & 86.5 ± 9.6 \\
    \hline
  \end{tabular}
  \caption{\label{tab:models_appendix} 
  Percentage of connected components that exceed the 90\% purity threshold, computed using ground-truth labels (CP~(GT)) and model predictions labels (CP~(Pred)), i.e using eq \eqref{eq:component-purity} with corresponding labels. The column MM reports the proportion of components where the predicted majority label matches the ground-truth majority label (see eq.~\eqref{eq:majority-match}). Results are averaged across all 1D lenses.}
\end{table*}

\end{document}